\documentclass[final,twocolumn,10pt,conference,a4paper]{IEEEtran}
\usepackage{t1enc}
\usepackage[utf8]{inputenc}
\usepackage{amsmath,amssymb,amsthm}
\usepackage{graphicx}
\usepackage{url}
\usepackage[pdftex,colorlinks=true]{hyperref}
\usepackage{extarrows}
\usepackage{algorithm}
\usepackage{algpseudocode}
\usepackage{balance}

\begin{document}
\IEEEoverridecommandlockouts
\title{Scheduling and Communication Schemes for Decentralized Federated Learning}
\author{%
  \IEEEauthorblockN{Bahaa-Eldin Ali Abdelghany\IEEEauthorrefmark{1},
    A. Fernández-Vilas\IEEEauthorrefmark{2},
    M. Fernández-Veiga\IEEEauthorrefmark{2},
   Nashwa El-Bendary\IEEEauthorrefmark{1},\\
    Ammar M. Hassan\IEEEauthorrefmark{1},
    Walid M. Abdelmoez\IEEEauthorrefmark{1}}
  \IEEEauthorblockA{\IEEEauthorrefmark{1}Arab Academy for Science, Technology \&
    Maritime Transport, Cairo, Egypt}%
  \IEEEauthorblockA{\IEEEauthorrefmark{2}atlanTTic, University of Vigo, Spain}%
  \thanks{This work was supported by the Spanish
    Government under research project ``Enhancing Communication
    Protocols with Machine Learning while Protecting Sensitive Data
    (COMPROMISE)'' PID2020-113795RB-C33, funded by
    MCIN/AEI/10.13039/501100011033. This work was partially done while Bahaa-Eldin Ali 
    was with the University of Vigo under EU-funded program KA-107.}}
\maketitle

\begin{abstract}
  Federated learning (FL) is a distributed machine learning paradigm in
  which a large number of clients coordinate with a central server to
  learn a model without sharing their own training data. One
  central server is not enough, due to problems of connectivity
  with clients. In this paper, a decentralized federated learning 
  (DFL) model  with the stochastic gradient descent (SGD) algorithm has been introduced,   as a more scalable approach to improve the learning performance in a network of agents   with arbitrary topology. Three scheduling  policies for DFL have been proposed for communications between the clients and the parallel servers, and the convergence, accuracy,  and loss have been tested  in a totally decentralized implementation of SGD. The  experimental results show that  the proposed scheduling polices have an impact both on the speed of convergence and in the final global model.

\end{abstract}


\section{Introduction}
\label{sec:intro}

Data generated at device terminals has recently increased exponentially, owing to the explosive growth of powerful individual computing devices worldwide and the rapid advancement of the Internet of Things (IoT). Data-driven machine learning is becoming a popular technique for making predictions and decisions about future events by making full use of massive amounts of data. Federated Learning (FL), a promising data-driven machine learning variant, provides a communication-efficient approach for processing large amounts of distributed data and is gaining popularity.FL was first proposed as a critical technique of distributed machine learning in a centralised form, in which edge clients perform local model training in parallel and a central server aggregates the trained model parameters from the edge without transmitting raw data from the edge clients to the central server.

FL was first tested on  a Google Android keyboard (Gboard). It
supports multilingual typing, including Google searches and sharing
results from the keyboard, as well as auto-correction, voice typing,
and glide typing. When Gboard displays some suggestions on the screen
based on user behavior, local learning occurs, and FL gains sway by
improving future suggestions/interactions with the user. As a result,
improved features such as next-word prediction, word completion,
corrections, and many more are available. To implement and experiment
FL on decentralized data, the following open-source frameworks are in
development/available: TensorFlow Federated (TFF)~\cite{tensorflow},
Federated AI Technology Enabler (FATE)~\cite{FederatedAI},
PySyft~\cite{PySyft}, PaddleFL~\cite{PaddleFL}, Clara Training
Framework~\cite{Nvidia}.

FL plays a crucial role in supporting the privacy protection of user
data and deploying in a complex environment with massive intelligent
terminal access to the network center due to the property of transmitting model parameters instead of user data and the distributed network structure that an arbitrary number of edge nodes are coordinated through one central server. FL operates as a centralized model with decentralized data, which makes the central server a critical point of failure.  Besides, in many cases of interest, not every client has a direct connection to the server for learning of the global model~\cite{Chen2020}.

Decentralized Federated Learning (DFL) is therefore well suited to implement
distributed learning with multiple data aggregators, thus reducing the cost of 
communication and the workload of the central server. DFL has been proposed and analyzed  in the literature under specific updating algorithms in the distributed servers (see~\cite{Yuan2016,Sirb2018,Xing2021}, but these works rely on a strict and rigid strategy  for the update and communication phases of the protocol. This might be difficult to achieve in general networks, where density and asymmetry are frequent. Consequently, the impact of the scheduling policy between the nodes acting as clients and the servers is a complex decision affecting the global behavior of learning algorithm.

In this work,  scheduling policies have been proposed for DFL designed
 to get insights about convergence, loss, and accuracy. The paper is organized as follows. In Section~\ref{sec:related-work}, the related work is introduced. Next, Section~\ref{sec:model} 
 presents the proposed DFL system model and the main assumptions. The experimental results are presented and discussed in  Section~\ref{sec:results}, and finally conclusions appear in Section~\ref{sec:conclusions}.

\section{Related Work}
\label{sec:related-work}

Numerous authors have discussed and created solutions for problems with FL resource allocation in their writings. The FL problem over wireless networks formulated in~\cite{Tran2019} captures the following trade-offs : (1) using the Pareto efficiency model, measuring learning time in relation to customer energy use, and (2) computation versus communication learning time by determining the ideal learning
accuracy. In~\cite{Shi2019}, by developing a joint bandwidth allocation and scheduling issue to reduce training time and achieve the desired model accuracy, the authors suggest a method for increasing the convergence rate of FL training concerning time. For the bandwidth allocation problem, they design an efficient binary search algorithm, while for maximum device scheduling, they adopt a greedy approach for achieving a trade-off between the latency and learning efficiency in each round. In~\cite{Chen2019}, The authors define the joint learning, wireless resource allocation, and client selection problem as an optimization problem to minimize the FL loss function. In~\cite{Khan2020}, the authors describe a method for self-organizing FL over wireless networks. They use a heuristic algorithm to minimize global FL time while taking local energy consumption and resource blocks into account.

IN \cite{Yang2020}, the authors suggests a paradigm for evaluating and
describing FL performance. For the convergence rate of FL, traceable
expressions are generated that consider the impact of inter-cell
interference as well as scheduling strategies. They also looked at the
efficiency (convergence rate) of scheduling rules such as proportional
fair scheduling, round robin scheduling, and random scheduling.
Other works have begun examining scheduling
strategies influenced by the chances for model improvement during FL
rounds. According to the channel circumstances and the importance of
local model updates, \cite{Amiri2020} establishes
scheduling policies for selecting the subset of devices to handle the
transmission inside each round. 

In wireless networks with clients sharing a single wireless link, the
contribution of~\cite{Xu2021} offers a long-term perspective for
resource allocation. The method is grounded in experimental
observation showing that choosing fewer customers during the initial
learning rounds and then gradually increasing this number is the
strategy having the best impact on learning performance. 
In~\cite{Chai2020}, the clients are divided into tiers according to
how well they performed during training, and an adaptive tier-based
client selection method is used in the authors' proposed Tier-based
Federated Learning (TEFL) System. In~\cite{Ren2020}, it is suggested a
scheduling strategy to make use of variation in multi-user channels as
well as diversity in the significance of edge device learning updates
(measured by gradient divergence). 
In~\cite{Huang2020}, the authors propose a proactive
algorithm that selects mobile clients based on predictions of their
future training and reporting abilities. The adopted approach is
divided into two parts: (1) In a metropolitan mobile edge computing
environment, predicting users' mobility trajectory patterns and the
apps they use on their smartphones, as well as (2) a deep
reinforcement learning-based client-selection algorithm handling
unanticipated dynamic events, are all possible. CPU, bandwidth, GPS
coordinates, and the success or failure of downloading and uploading
local and global parameters are the metrics that are observed and
forecast. 

The authors at~\cite{FedAvg} advocate the decentralized approach
that leaves the training data distributed on the edge devices and
learns a shared model by aggregating locally computed updates. They
show a practical method for FL of deep networks based
on iterative model averaging, and conduct an extensive empirical
evaluation, considering five different model architectures and four
datasets. The Decentralized SGD is a driving
engine for FL and its performance is influenced by
internode communications and local updates. At~\cite{Liu2022} propose
a DFL framework that implements both
internode communication periodically and multiple local updates to
strike a balance between communication efficiency and model
consensus. They establish strong convergence guarantees for the DFL
algorithm without the assumption of convex objects. 

\section{Model of Peer-to-Peer Decentralized FL}
\label{sec:model}

\begin{figure}
  \centering
  \includegraphics[width=0.8\columnwidth]{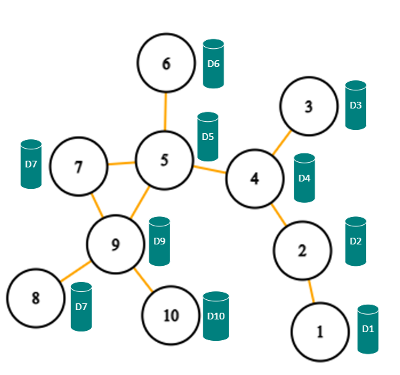}
  \caption{\label{fig:network-graph} System Model for Decentralized
    Federated Learning.} 
\end{figure}
\begin{table*}[t]
  \centering
  \caption{\label{table:schedulers} Scheduling policies for the
    decentralized learning algorithm.}
  \begin{tabular}{lccc}
    & \textsc{scheduler a} & \textsc{scheduler b} &
                                                      \textsc{scheduler
                                                      c} \\ \hline
    Rounds $t = 2k + 1$ & $\{2, 3 \} \to 4$, $\{6, 7\} \to 5$, $\{8,
                          10 \} \to 9$ & $\{2, 3 \} \to 4$, $\{ 6 \}
                                         \to 5$, $\{7, 8, 10 \} \to 9$
    & $\{2, 3 \} \to 4$, $\{8, 10 \} \to 9$ \\

    Rounds $t = 2k$ & $\{4, 9 \} \to 5$, $\{ 1 \} \to 2$
                            & $\{4, 9 \} \to 5$, $\{ 1 \} \to 2$
    & $\{4, 6, 7, 9 \} \to 5$, $\{ 1 \} \to 2$ \\ \hline
  \end{tabular}
\end{table*}

The graph, the scheduler, and the global parameters aggregation are the three
parts of the system model that are defined in this section.the first part
explains how edge devices in the graph are connected to one
another. the second scheduler plans the aggregators and clients.
the third part is the aggregator way will combine the parameters given in the aggregation
section as well as the specifics of how the new global model
parameters will be calculated using FedAvg.

\subsection{Graph}

We consider a community of learning nodes modeled as an undirected
graph $G = (V, E)$ with $n = |V|$ nodes. For $i, j \in V$, edge
$(i, j) \in E$ represents a bidirectional communication link between
nodes $i$ and $j$. Each node is assumed to have access to a local
dataset $D_i, i \in V$, containing $d_i = |Di |$ samples of a common
unknown distribution D from which we are interested in learning, i.e.,
in building a statistical model $F_{\boldsymbol{\theta}}$ from a given
class, where $\boldsymbol\theta$ denotes the model parameters which
are to be optimized during the learning process. For instance,
$F_{\boldsymbol{\theta}}$ can be a (deep) neural network, and
$\boldsymbol\theta$ the weights between its adjacent layers. Different
from centralized and federated learning (FL), where only a single node
is in charge of building $F_{\boldsymbol{\theta}}$, in decentralized
or distributed learning (DL) we allow each node to learn from its
neighbors, possibly in an asynchronous way.

\subsection{Schedulers}

The scheduler role is choosing which node will work as an aggregator
for the parameters from specific neighbors. We propose, for the graph
depicted in Figure~\ref{fig:network-graph}, the three scheduling
policies listed in Table~\ref{table:schedulers}. The notation
$\{X, Y \} \to Z$ is used to denote that nodes $X$ and $Y$ are clients
and node Z is aggregation node. The criteria for choosing a node
to work as an aggregator is based on the round number and it is
expressed by the formula Rounds $t = 2k + 1$ for odd rounds, Rounds
$t = 2k$ for even rounds.

We see at Table~\ref{table:schedulers}, for instance, that node $5$ aggregates weights each round in
scheduler A. This node was crucial in this scheduler since it
aggregates from the other aggregators, so it will hold the most recent
model that was averaged across multiple nodes.in every scheduler, the sequence of communications between clients and servers yields, after a pair of rounds, a connected graph.

\subsection{Decentralized Federated Averaging}
\label{sec:description}

\begin{figure*}
  \centering
  \includegraphics[width=0.90\textwidth]{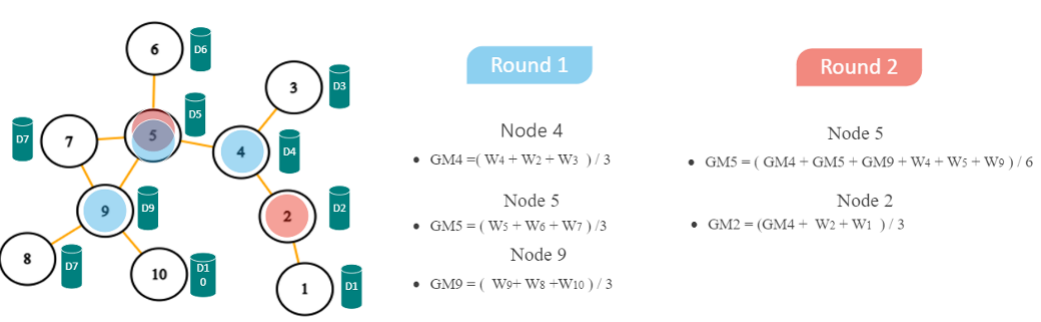}
  \caption{\label{fig:tracking}Tracking Models for first two rounds for Scheduler A.}
\end{figure*}

The FedAvg Algorithm is the most widely used technique for calculating the global model parameters. We anticipate that the aggregators' average model will satisfy full convergence. McMahan~\cite{FedAvg} proposed FedAvg, in which clients collaboratively send updates of locally trained models to the aggregator node, each client running a local copy of the global model on its local trainning data. The global model weights are then updated with an average of the updates from clients and deployed back to clients. This extends previous training work by not only supplying local models but also performing training on each device locally. As a result, FedAvg may enable clients (particularly those with small datasets) to collaboratively learn a shared prediction model while retaining all training data locally. before aggregators construct the global model. neighbors send new model parameters and old global parameters that were kept from previous rounds. The aggregator sums new parameters and old global parameters from itself and the received from neighbors ,then calculate the average to build the new global parameters by using FedAvg. If any node did not participate in any previous rounds, it will send only new local parameters to the aggregator. as it will not have old global parameters. Fig~\ref{fig:tracking} Aggregator nodes are coloured blue in the first round and red in the second. In the first and second rounds, we concentrate on node 5. In the first round, node 5 will aggregate weights from neighbors as well as the old global model four GM4. Node 4 created GM4 while acting as an aggregator. After which a new global model was calculated The new global model is known as GM5, which stands for Global model for node 5. In the second round of GM5, weights from neighbors are added to the old global models from previous rounds. The total will be divided by the number of models. The same criteria will be used in subsequent odd and even rounds.

\label{sec:implementation}

Pseudo code~\ref{alg:peer-to-peer-FedAvg} for each round there are one aggregator and set of clients. When the round starts the aggregator will initialize the clients with random weights . clients will get the random weights or last round weights  that are kept from the aggregator. clients and their aggregator will start executing local computations for epochs. the updates will be sent to the aggregator. The aggregator will combine clients updates with its update then generate new parameters using FedAvg. The new parameters will be sent to clients and each client will save it locally . Clients will use global parameters to initiate their clients when they work as aggregators.after round finished the algorithm extracts a new aggregator. the aggregators determined by schedulers.

\begin{algorithm}[t]
  \caption{\label{alg:peer-to-peer-FedAvg} Decentralized FedAVg. }
  \begin{algorithmic}[1]
    \While {$Q$ is not empty}

    \For {each round  $t = 1,2 \dots $}
        \State $AG \leftarrow \text{Dequeue(Q) }$ \Comment{extract a new node to work as aggregator from queue}

        \State $N \leftarrow \text {set of neighbors/clients of Aggregator (AG)}$
    \If {$w_{t-1}$ Exist}
    
    \State $\mathbf{w}_t \leftarrow \mathbf{w}_{t-1}$\Comment{set initial weights with last round weights aggregated}

   \Else
   \State $\mathbf{w}_t \leftarrow \mathbf{w}_{0}$\Comment{set initial weights randomly}
   \EndIf
    \State $AddNeighbour(N,AG)$ \Comment{add aggregator node to neighbor set}

    \For {each neighbour  $X$ in $N$}
\State $\mathbf{w}^d_{t+1} \leftarrow \mathsf{NeighborUpdate}(d, \mathbf{w}_t)$    
    \State $\mathbf{w}_{t+1} \leftarrow \sum_{i = 1} ^{|N_x|} \frac{n_i}{n} \mathbf{w}_{t+1}^k$
\EndFor
    \EndFor
    \EndWhile
  \end{algorithmic}
\end{algorithm}

\section{Experimental Results}
\label{sec:results}

\begin{figure*}[p]
  \centering
  \includegraphics[width=0.85\textwidth]{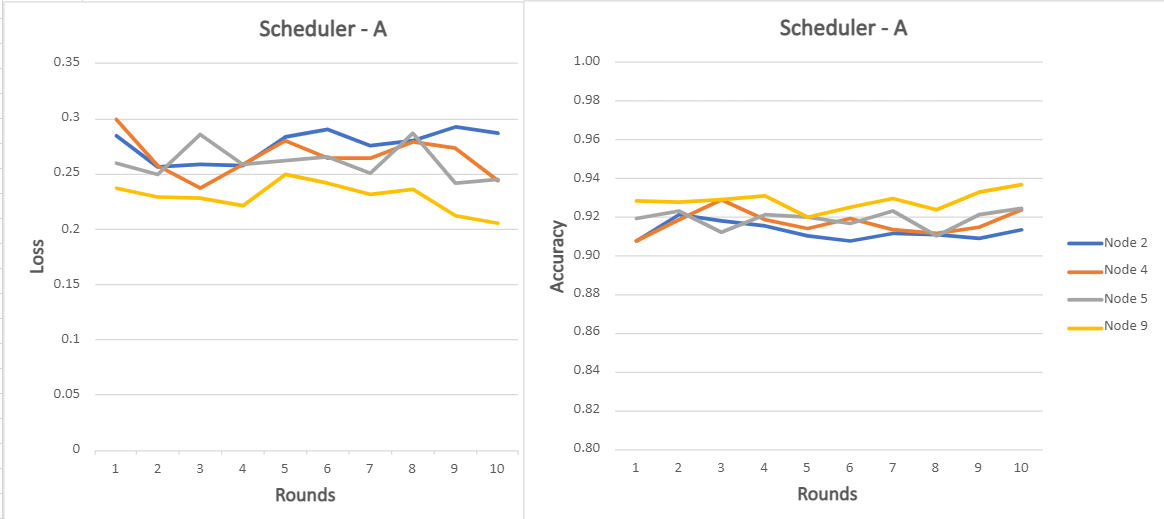}
  \includegraphics[width=0.85\textwidth]{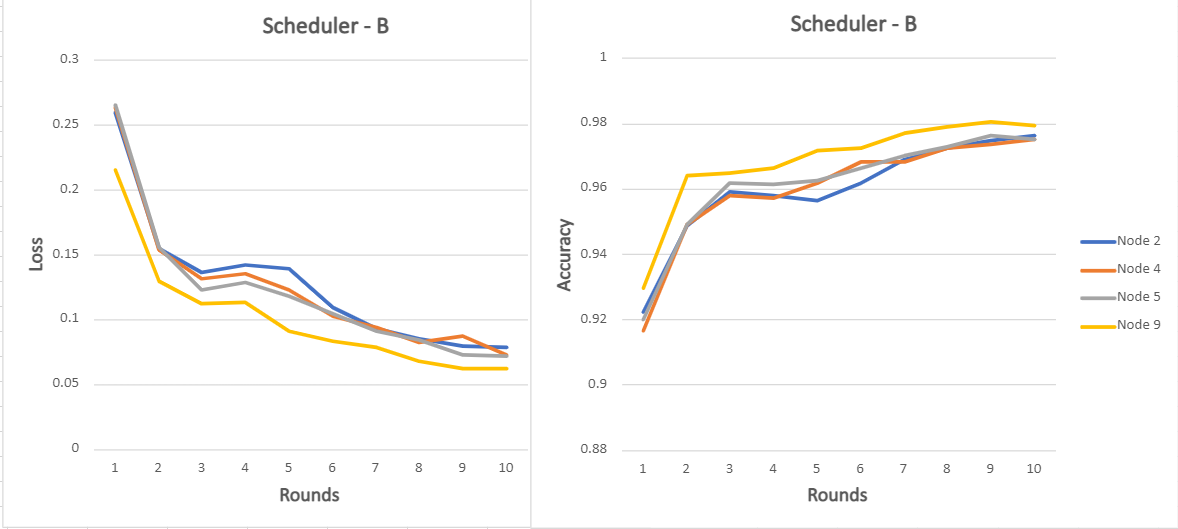} 
  \includegraphics[width=0.85\textwidth]{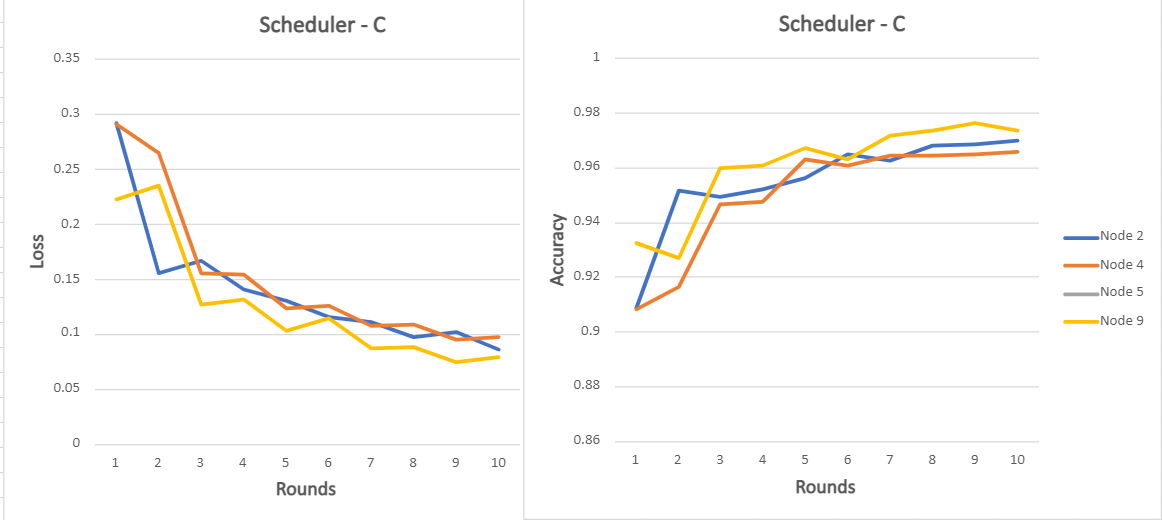}

  \caption{\label{fig:server1} Performance Across Schedulers and Nodes
  in the graph.}
\end{figure*}

\subsection{Environment}

\subsubsection{Flwr}

The implementation was developed under an Flwr environment with
anaconda with python. The Project itself contains one python
code. When this code is executed the graph description is
read. working on Flwr gives the flexibility to simulate real-world
scenarios as we need to simulate the communication noise between
nodes, and it can work in simulation mode which is very good at
simulating nodes to get results quickly.

\subsubsection{Google Colab}

Google-colab is an online website that we can use to run machine
learning model notebooks in the cloud. It provides a good GPU and CPU
to run ML notebooks smoothly and get results and save outputs so
easily.

\subsubsection{WandB}

WandB is abbreviated for weight and bias~\cite{Wandb} it is an online
API that is used with our ML Project for teaching and logging results
in an interactive way. It is a very helpful tool for good
visualizations and creating reports and notes for any ML Model and
tracking parameters, supports many formats and representations to
download or view results.

\subsubsection{MNIST}
The Modified National Institute of Standards and Technology for handwritten digits 
was used in our experiment to assess the proposed Decentralized 
FL.

\subsection{Results}

\begin{table*}[t]
    \caption{\label{table:resultsAB} Loss and accuracy for the network graph and scheduler A and B, per round of learning.}
    \begin{center}
    \begin{tabular}{ccccc|cccc} \hline
    & \multicolumn{4}{c}{\textsc{scheduler a}} & \multicolumn{4}{c}{\textsc{scheduler b}} \\
    \textsc{round} & \textsc{node 2} & \textsc{node 4} & \textsc{node 5} & 
    \textsc{node 9} & \textsc{node 2} & \textsc{node 4} & \textsc{node 5} & 
    \textsc{node 9} \\ \hline
    $1$ & $(0.29, 91\%)$ & $(0.3, 92\%)$ & $(0.26, 92\%)$ & $(0.24, 93\%)$ 
    & $(0.26, 92\%)$ &	$(0.26, 92\%)$ & $(0,27, 92\%)$ & $(0.22, 93\%)$ \\
    $2$	& $\mathbf{(0.26, 92\%)}$ & $(0.26, 92\%)$ & $\mathbf{(0.25, 92\%)}$ & $(0.23, 93\%)$ 
    & $(0.15, 95\%)$ & $(0.15, 95\%)$ & $(0.16, 95\%)$ & $(0.13, 96\%)$ \\
    $3$	& $\mathbf{(0.26, 92\%)}$ & $\mathbf{(0.24, 93\%)}$ & $(0.29, 91\%)$ & $(0.23, 93\%)$ 
    & $(0.14, 96\%)$ & $(0.13, 96\%)$ & $(0.12, 96\%)$ & $(0.11, 97\%)$ \\
    $4$	& $\mathbf{(0.26, 92\%)}$ & $(0.26, 92\%)$ & $(0.26, 92\%)$ & $(0.22, 93\%)$ 
    & $(0.14, 96\%)$ & $(0.14, 96\%)$ & $(0.13, 96\%)$ & $(0.11, 97\%)$ \\
    $5$	& $(0.28, 91\%)$ & $(0.28, 91\%)$ & $(0.26, 92\%)$ & $(0.25, 92\%)$ 
    & $(0.14, 96\%)$ & $(0.12, 96\%)$ & $(0.12, 96\%)$ & $(0.09, 97\%)$ \\
    $6$ & $(0.29, 91\%)$ & $(0.26, 92\%)$ & $(0.27, 92\%)$ & $(0.24, 93\%)$ 
    & $(0.11, 96\%)$ & $(0.1, 97\%)$ & $(0.1, 97\%)$ & $(0.08, 97\%)$ \\
    $7$	& $(0.28, 91\%)$ & $(0.26, 91\%)$ & $\mathbf{(0.25, 92\%)}$ & $(0.23, 93\%)$ 
    & $(0.09, 97\%)$ & $(0.09, 97\%)$ &	$(0.09, 97\%)$ & $(0.08, 98\%)$ \\
    $8$ & $(0.28, 91\%)$ & $(0.28, 91\%)$ & $(0.29, 91\%)$ & $(0.24, 92\%)$ 
    & $(0.09, 97\%)$ & $(0.08, 97\%)$ & $(0.08, 97\%)$ & $(0.07, 98\%)$ \\
    $9$ & $(0.29, 91\%)$ & $(0.27, 92\%)$ & $(0.24, 92\%)$ & $(0.21, 93\%)$ 
    & $(0.08, 97\%)$ & $(0.09, 97\%)$ &	$\mathbf{(0.07, 98\%)}$ & $\mathbf{(0.06, 98\%)}$ \\
    $10$ & $(0.29, 91\%)$ & $(0.24, 92\%)$ & $\mathbf{(0.25, 92\%)}$ & $\mathbf{(0.21, 94\%)}$ 
    & $\mathbf{(0.08, 98\%)}$ & $\mathbf{(0.07, 98\%)}$ & $\mathbf{(0.07, 98\%)}$ & $\mathbf{(0.06, 98\%)}$ \\ \hline
    \end{tabular}   
    \end{center}
\end{table*}
\begin{table*}[t]
    \caption{\label{table:resultsC} Loss and accuracy for the network graph and scheduler C, per round of learning.}
    \begin{center}
    \begin{tabular}{ccccc} \hline
    \textsc{round} & \textsc{node 2} & \textsc{node 4} & \textsc{node 5} & 
    \textsc{node 9}  \\ \hline
    $1$ & $(0.29, 91\%)$ & $(0.29, 91\%)$ & --- & $(0.22, 93\%)$ \\
    $2$ & $(0.16, 95\%)$ &	$(0.27, 92\%)$ & $(0.28, 91\%)$ & $(0.23, 93\%)$ \\
    $3$ & $(0.17, 95\%)$ & $(0.16, 95\%)$ & --- & $(0.13, 96\%)$ \\
    $4$	& $(0.14, 95\%)$ & $(0.15, 95\%)$ & $(0.16, 95\%)$ & $(0.13, 96\%)$ \\
    $5$ & $(0.13, 96\%)$ & $(0.12, 96\%)$ & --- & $(0.1, 97\%)$ \\
    $6$	& $(0.12, 97\%)$ & $(0.13, 96\%)$ & $(0.13, 96\%)$ & $(0.11, 96\%)$ \\
    $7$	& $(0.11, 96\%)$ & $(0.11, 96\%)$ & --- & $(0.09, 97\%)$ \\
    $8$ & $(0.1, 97\%)$ & $(0.11, 96\%)$ & $(0.11, 96\%)$ &	$(0.09, 97\%)$ \\
    $9$ & $(0.1, 97\%)$ & $(0.1, 96\%)$ & --- &	$\mathbf{(0.08, 98\%)}$ \\
    $10$ & $\mathbf{(0.09, 97\%)}$ & $\mathbf{(0.1, 97\%)}$ &	$\mathbf{(0.1, 97\%)}$ &	$(0.08, 97\%)$ \\ \hline
    \end{tabular}   
    \end{center}    
\end{table*}    
    
\cite{Beutel2020}  has many advantages like stability and the capability of operating in real federated learning scenarios. Tracking both the loss and accuracy obtained for aggregator nodes in the graph is considered.
Tables~\ref{table:resultsAB} and~\ref{table:resultsC}, and Figure~\ref{fig:server1}
demonstrate the accuracy and loss through the rounds.  Regarding  these results, we can make the following observations.

\subsubsection{Scheduler A} In respect of aggregator nodes 4, 5, 2 and 9.
the first three rounds, node 4 loss is diminishing. before the last round, between 0.25 and 0.28 The lowest loss in the previous round was 0.24. Node 5 was 0.25 in the first round. Node 5 and subsequent rounds are unstable. It fluctuated between growing and falling, but it eventually fell to 0.24.node 2 at first round loss is 0.28 and accuracy is 91\%. rounds 2 ,3 and 4 same loss 0.25 for node 2 lower than first round.rounds through 5 - 9 loss in range 0.28 - 0.29. final round loss for node 2 is 0.29 and accuracy is 91\%.  
In the first round, node 9 has a value of 0.23. Except for rounds six, seven, and eight, loss dropped across the rounds. Loss is 0.24, 0.23, and 0.23, respectively. The last round had the lowest loss of 0.2. At final rounds aggregator accuracy is 92\%, whereas node 9 accuracy is 94\%. When compared to other aggregators or clients, Node 9 has the highest accuracy and lowest loss.

\subsubsection{Scheduler B} By monitoring node learning rates It is obvious that every node learning loss will decrease. This means that nodes are growing better with each cycle. Except for node 1, which participates only in even rounds, all nodes lose between 0.21 and 0.26 after the first round. The loss for node 1 in the second round is 0.16. Node 2 has a value of 0.15 while node 9 has a value of 0.12. Because node 2 and node 9 participated in the first round, their losses were lower than node 1. Through the rounds, all nodes lost less. The final round loss for nodes 3, 4, and 5 is 0.07. The loss at node 1 is 0.08. Node 9 has the lowest loss value of 0.06.
Note : node 1 not drawn at the chart.at final round, nodes 2,4,5 and 9 accuracies is 97\%. 

\subsubsection{Scheduler C} The first loss for aggregators 2, 4, 5, and 9 is 0.29, 0.29, 0.27, and 0.22, respectively. Node 2 loss decreases in all rounds except round nine, where it increases by 0.01. In the final round, node 2 had the lowest loss of 0.08. Node 4 loss is reduced through all rounds, with the exception of the last round, which grew by 0.002 and became 0.097. Even though they lost in the last round, they are still doing well. Except for round nine, node 9 loss is reduced across all rounds. The final round defeat is 0.079 percent more than the previous round loss. Throughout all rounds, the loss of Node 5 is reduced. The rate of change of loss at node 5 is large between rounds, as seen. by looking at the nodes in the network as a whole. Every round, the loss changed and dropped efficiently. For the 10 rounds, the accuracy of all nodes rose by about 7\%. They began with an accuracy of roughly 90\% and gradually grew to 97\%. When compared to other aggregators or clients, Node 9 has the highest accuracy and lowest loss.

\section{Conclusions and Future Work}
\label{sec:conclusions}
 
 Performance in FL depends crucially on whether full or partial participation from the nodes In this paper, thorough a series of case studies, we have shown that the design of the
 scheduling strategies between clients and servers ---where a node can act in different rounds either as a 
 client sending its own model/parameters or as an aggregator for a subset of its nearby neighbors--- it is fundamental 
 to balance the trade-off between the precision in the global model, the global learning rate, and the 
 communication costs. We have shown that, with different schedulers, not only convergence to a global model 
 is not substantially affected by the network topology, but that instead the rate of learning at the
 different nodes in the graph is rather homogeneous, despite slight stochastic variations due to the local
 topology. Nevertheless, even though the global model retains a similar quality as compared to
 centralized FL, in DFL the scheduling of clients and servers turns out to be essential for minimizing 
 the number of messages exchanged, and also for guaranteeing that all nodes in the system learn the global 
 model at a similar pace. The highest accuracy shown by the proposed model is 97\%, and the lowest loss is 0.07.
 A few promising future directions include measuring communication cost the number of messages exchanged.

\balance
\bibliographystyle{IEEEtran}
\bibliography{DFL}

\end{document}